\pdfoutput=1

\documentclass[11pt]{article}

\usepackage{coling}

\usepackage{times}
\usepackage{latexsym}
\usepackage[utf8]{inputenc} 
\usepackage[T1]{fontenc}    
\usepackage{hyperref}       
\usepackage{url}            
\usepackage{booktabs}       
\usepackage{amsfonts}       
\usepackage{nicefrac}       
\usepackage{microtype}      
\usepackage{xcolor}         
\usepackage[ruled,vlined]{algorithm2e}
\usepackage{graphicx}
\usepackage{soul}
\usepackage{tikz}
\usepackage{amssymb}
\usepackage{lipsum}
\usepackage{wrapfig}
\usepackage{caption}
\usepackage{mathtools}
\usepackage{multirow}
\usepackage{algpseudocode}
\usepackage{comment}
\usepackage{subcaption}
\usepackage{amsmath}
\usepackage{enumitem}
\setlist[enumerate]{itemsep=0mm}
\setlist{nosep}

\newcommand{\turbo}{{\texttt{GPT3.5Turbo}}\xspace}

\newcommand{\gptt}{{\texttt{GPT-4Turbo}}\xspace}
\newcommand{\mix}{{\texttt{Mixtral}}\xspace}
\newcommand{\indic}{{\texttt{IndicQA}}\xspace}
\newcommand{\tydi}{{\texttt{TyDiQA}}\xspace}
\newcommand{\squad}{{\texttt{SQuAD-F1}}\xspace}
\newcommand{\meta}{{\texttt{MLQA-F1}}\xspace}
\newcommand{\mono}{{\texttt{Mono}}\xspace}
\newcommand{\trans}{{\texttt{Trans}}\xspace}
\newcommand{\siml}{{\texttt{Sim}}\xspace}
\newcommand{\aggsrc}{{\texttt{Agg\_Src}}\xspace}
\newcommand{\aggtrans}{{\texttt{Agg\_Trans}}\xspace}
\newcommand{\gpteval}{{\texttt{GPTAnnotator}}\xspace}
\newcommand{\gptevalf}{{\texttt{GPTAnnotator-F1}}\xspace}
\newcommand{\haf}{{\texttt{HumanAnnotator-Score}}\xspace}

\title{Bridging the Language Gap: Dynamic Learning Strategies for Improving Multilingual Performance in LLMs}

\newcommand*\samethanks[1][\value{footnote}]{\footnotemark[#1]}

\author{%
Somnath Kumar\thanks{Equal Contributions} \quad Vaibhav Balloli\samethanks \quad Mercy Ranjit \quad Kabir Ahuja \quad \\ \bf Sunayana Sitaram \quad Kalika Bali \quad Tanuja Ganu \quad Akshay Nambi \\
\bf Microsoft Research India \\
\texttt{akshayn@microsoft.com}
}

\begin{document}

\maketitle

\begin{abstract}
Large language models (LLMs) have revolutionized various domains but still struggle with non-Latin scripts and low-resource languages.  This paper addresses the critical challenge of improving multilingual performance without extensive fine-tuning. We introduce a novel dynamic learning approach that optimizes prompt strategy, embedding model, and LLM per query at runtime. By adapting configurations dynamically, our method achieves significant improvements over static, best and random baselines. It operates efficiently in both offline and online settings, generalizing seamlessly across new languages and datasets. Leveraging Retrieval-Augmented Generation (RAG) with state-of-the-art multilingual embeddings, we achieve superior task performance across diverse linguistic contexts. Through systematic investigation and evaluation across 18 diverse languages using popular question-answering (QA) datasets we show our approach results in 10-15\% improvements in multilingual performance over pre-trained models and 4x gains compared to fine-tuned, language-specific models. 

\end{abstract}

\vspace{-10pt}
\section{Introduction \& Related Work}
\label{sec:introduction}
Large Language Models (LLMs), such as ChatGPT~\cite{openai2023gpt4}, Gemini~\cite{team2023gemini}, and Claude~\cite{anthropic2023claude3}, have driven significant advancements in artificial intelligence (AI), setting new benchmarks for performance across a wide range of tasks~\cite{DBLP:journals/corr/abs-2005-14165, ouyang2022training, openai2023gpt4}.  They excel in diverse applications, including search engines, office tools, and critical sectors like health, education, and agriculture~\cite{shiksha, farmerchat, khanmigo, m365copilot}. By transforming workflows, LLMs are rapidly becoming essential in real-world systems, revolutionizing approaches to complex tasks across domains.

However, despite their widespread success, LLMs remain predominantly optimized for English and Latin-script languages, creating significant limitations in non-English and multilingual environments~\cite{mega, ahuja2023megaverse, khanuja2021muril}. 

\vspace{-5pt}
\begin{wraptable}{r}{5cm}
\resizebox{0.3\textwidth}{!}{%
\begin{tabular}{ll}
\hline
Method              & Accuracy                    \\ \hline
LLama2 70B          & {8.5}  \\
Mistral 7B instruct & {29.6} \\ \hline
Cohere              & {78.8} \\
Palm2               & {76.5} \\
GPT3.5              & 60.1                        \\
GPT4                & 71.5                        \\ \hline
TULR-XXL            & \textbf{84.6}      \\ \hline        
\end{tabular}%
}\vspace{-5pt}
\caption{Performance comparison across various models for \tydi.}
\label{tab:motiv}
\vspace{-8pt}
\end{wraptable}
Although efforts have been made to extend LLM capabilities to low-resource languages through fine-tuning and smaller, specialized models~\cite{gala2024airavata, abdin2024phi}, their performance in multilingual tasks still lags behind state-of-the-art (SOTA) multilingual models like TULRv6 and XLMR~\cite{goyal2021larger}. A comparative analysis, shown in Table~\ref{tab:motiv}, highlights this performance gap across LLMs such as GPT-3.5, GPT-4, Palm2, and LLaMA2 on the \tydi multilingual QA dataset, where they consistently underperform relative to models specifically designed for multilingual tasks.

To bridge the performance gap in multilingual LLMs, two key research directions have emerged~\cite{qin2024multilingual, huang2024survey}. The first focuses on enhancing foundational models with additional multilingual data, such as Cohere AI's Aya 101~\cite{cohereaya}, which curates instructions across 99 languages. However, this approach has limitations. Data scarcity for low-resource languages remains critical, leading to suboptimal performance during pre-training~\cite{hämmerl2022multilingual, DBLP:journals/corr/abs-2004-06748}. Additionally, the computational cost of training models across multiple languages is prohibitive, making fine-tuning impractical for many researchers~\cite{qin2024multilingual, liu2024understanding}. Even after fine-tuning, models often struggle to generalize beyond the languages or tasks they were trained on, as seen in Sarvam 2B~\cite{sarvam, sarvamai2024openhathi,xu2024survey}, which underperforms on Indian languages excluded from its training data.

The second direction aims to improve pre-trained LLMs through optimized external configurations, focusing on \textbf{(1) Prompt Strategies}, \textbf{(2) Embedding Models}, and \textbf{(3) Model Selection}. While various prompt strategies (e.g., Chain-of-Thought, cross-lingual prompts) have improved specific tasks~\cite{weichain,shi2022language,prompt1,li2024eliciting}, no single approach works consistently across all languages~\cite{zhao2021discrete,huang2022zero, fu2022polyglot,lin2021few}. For instance, Chain-of-Thought prompting improves reasoning in English~\cite{weichain,lai2024mcot} but struggles with languages like Finnish or Tamil.

Embedding models like OpenAI’s text-embedding-3 and Cohere's multilingual v3.0~\cite{openai2024ada3, cohere2024embedv3} have significantly boosted multilingual performance in question-answering tasks, yet selecting the right embedding remains challenging as performance varies across languages. Furthermore, the release of new LLMs exacerbates the \textbf{Model Selection Dilemma}, where model performance varies widely across languages and retraining models for each version is impractical due to resource constraints.

Most prior work relies on \textbf{static configurations}—where a single prompt strategy or embedding is applied to specific tasks~\cite{qin2024multilingual, huang2024survey}. This falls short in multilingual contexts due to linguistic diversity. For example, strategies optimized for English may fail in languages like Japanese or Arabic, while embeddings designed for Indo-European languages may struggle with tonal languages like Mandarin~\cite{mega,ahuja2023megaverse}. These challenges emphasize the need for real-time, dynamic approaches that adapt to each language's unique requirements without requiring costly retraining. This is crucial in multilingual settings where a one-size-fits-all configuration is unlikely to succeed. 

Our work addresses this gap with a dynamic runtime selection framework. Unlike static configurations, our approach dynamically selects the best combination of prompt, model, and embedding for each query based on the task and language. For instance, a French query may use a model fine-tuned for Western European languages and a prompt strategy that handles gendered nouns, while a Hindi query might employ a strategy suited for free word order and compound verbs. This real-time adaptability ensures each query receives an accurate, context-aware response tailored to its linguistic structure.

Our key contributions are twofold:
\begin{enumerate}
   \itemsep0em 
 \item \textbf{Hybrid Approach:} We integrate LLM-generated responses with multilingual embeddings in a Retrieval-Augmented Generation (RAG) setup. This hybrid model improves document retrieval and text generation, enhancing coherence and relevance, while addressing performance gaps in multilingual tasks. By using language-specific embeddings, we bridge cross-lingual understanding, achieving superior results, especially in low-resource languages.
 \item \textbf{Dynamic Learning Framework:} We introduce a dynamic configuration framework that optimizes runtime selection of prompts, LLMs, and embeddings. Powered by a lightweight transformer, this framework generalizes across tasks, languages, and datasets without retraining for each domain, reducing computational overhead. By selecting optimal configurations in real time, it ensures adaptability to new LLMs, embedding models, or prompt strategies as they emerge.
\end{enumerate}

Our hybrid dynamic learning architecture combines LLMs with convolutional layers, supporting both offline and online learning. It addresses three key needs: \textbf{(i) Offline Learning}, leveraging ground-truth data for optimal configuration in controlled settings; \textbf{(ii) Online Adaptability}, adjusting dynamically to real-time inputs and distribution shifts; and \textbf{(iii) Language and Dataset Flexibility}, maintaining high performance across diverse linguistic and contextual variations.

We validate our approach using the \indic and \tydi QA datasets, which encompass 18 languages. Our framework demonstrates a 10-15\% improvement in multilingual performance compared to existing pre-trained LLMs, and significantly outperforms fine-tuned models optimized for specific languages, such as Ambari~\cite{Ambari}, Airavata~\cite{Airavata}, and Navarasa~\cite{Navarasa}, with performance gains exceeding 4x. These results demonstrate the superiority of our dynamic approach, which outperforms both static models and fine-tuned, language-specific solutions.

\begin{figure*}[!ht]
\vspace{-5pt}
\begin{minipage}[l]{0.5\linewidth}
        \includegraphics[width=0.95\columnwidth]{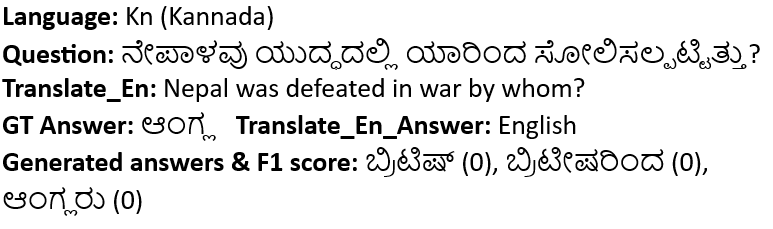}
\end{minipage}
\begin{minipage}[l]{0.5\linewidth}
       \includegraphics[width=0.95\columnwidth]{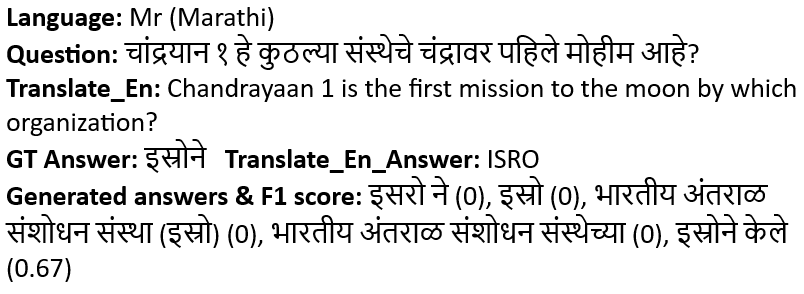}
\end{minipage}
\vspace{-5pt}
 \caption{Examples showing the limitations in the GT answer in \indic dataset.}
        \label{fig:examples}
        \vspace{-18pt}
\end{figure*}
While we evaluate on QA tasks, our dynamic framework is versatile and extends to other multilingual applications. By decoupling task performance from any single model, prompt, or embedding, it provides an efficient, scalable solution for overcoming LLM limitations in non-English and low-resource languages. 

\section{Multilingual Tasks, Datasets \& their Limitations}
\label{sec:setup}
In this work, we focus on RAG-based Question Answering (QA) tasks, demonstrating the model's ability to deliver accurate responses by leveraging external text context.
\subsection{Dataset}
\label{sec:dataset}
\begin{wraptable}{r}{4cm}
\vspace{-10pt}
\caption{Datasets}
\label{tab:dataset}
\scalebox{0.7}{
\begin{tabular}{ll|ll}
\toprule
\multicolumn{2}{l}{\textbf{IndicQA}} & \multicolumn{2}{l}{\textbf{TyDiQA}} \\ \toprule
Lang         & \# Q          & Lang         & \# Q         \\ \toprule
as           & 1789         & bn           & 180         \\
bn           & 1763         & te           & 874         \\
gu           & 2017         & fi           & 1031        \\
hi           & 1547         & ko           & 414         \\
kn           & 1517         & ru           & 1079        \\
ml           & 1589         & ar           & 1314        \\
mr           & 1604         & en           & 654         \\
or           & 1680         & id           & 773         \\
pa           & 1542         & sw           & 596         \\
ta           & 1804         &              &             \\
te           & 1734         &              &       \\  \bottomrule     
\end{tabular}
}
\vspace{-15pt}
\end{wraptable}

\vspace{-5pt}
We utilize two prominent multilingual QA datasets that includes 18 diverse languages (see Table~\ref{tab:dataset}) from high to medium to low resource including Latin and Non-Latin scripts (we follow ISO 639-1 language code standards in the remaining of the paper~\cite{iso}):

1.\indic~\cite{ai4bharat2022indicqa}: A curated dataset in 11 Indic languages sourced from Wikipedia on topics related to Indic culture and history, comprising over 18,000 questions. 

2.\tydi~\cite{tydiqa}: This dataset covers 9 typologically diverse languages. Our experiments focus on the Gold-P task, where only the gold answer passage is provided rather than the entire Wikipedia article. 
\vspace{-5pt}
\subsection{Evaluation Metrics for Multilingual QA}
\label{subsec:eval_metrics}
\vspace{-5pt}
F1 score is the commonly used metric in QA tasks~\cite{DBLP:journals/corr/RajpurkarZLL16}, compares individual words in predictions to the True Answer. While \squad is standard for English QA evaluation, \meta~\cite{DBLP:journals/corr/abs-1910-07475} offers additional preprocessing for fair multilingual evaluation, including stripping Unicode punctuations and stand-alone articles. Hence, we adopt \meta as our evaluation metric. 


\subsection{Limitations of Current Datasets \& Evaluation Approach}
\label{sec:limitations}
Many multilingual evaluation datasets were developed before the advent of Large Language Models (LLMs), posing two key challenges:

\texttt{Challenge 1: Limited Ground Truth (GT).} These datasets usually contain only one answer per question, though multiple semantically correct answers may exist, particularly in real-world and conversational contexts.\\
\texttt{Challenge 2: Strict Evaluation Metrics.} The standard F1 scoring at the word level leads to significant penalties for minor variations in answers, especially when only a single GT is available.

Figure~\ref{fig:examples} demonstrates these challenges using the \indic dataset for Kannada and Marathi. Although generated responses are factually correct, they differ slightly from the single GT answer, resulting in low or zero \meta scores. This underscores the limitations of both the dataset's GT and the evaluation method.

One solution is to enrich GTs by including all valid alternatives, but this requires extensive and costly data collection. To overcome this, we introduce \gpteval, leveraging an LLM (e.g., GPT-4) to validate predicted answers. This builds on previous work where GPT models are used as evaluators and annotators for diverse tasks. \gpteval assesses predicted responses by comparing them to the original GT and outputs three options: \textit{YES} for semantically correct answers, \textit{NO} for mismatches, and \textit{PARTIAL} for partial matches. \gpteval enriches the GT with correct answers, creating a more comprehensive reference set (see Appendix~\ref{sec:gptanno} for prompts).

To further refine evaluations, we propose \gptevalf, an F1 score calculated against the enriched GT with multiple valid answers. In contrast, the traditional \meta score compares predictions against the original, limited GT. Both are F1 metrics but differ in the number of answers they evaluate against (\meta uses a single-answer GT, while \gptevalf considers multiple correct answers).

\begin{wrapfigure}{r}{4.7cm}
    \centering
\includegraphics[width=0.28\textwidth, height=1.2in]{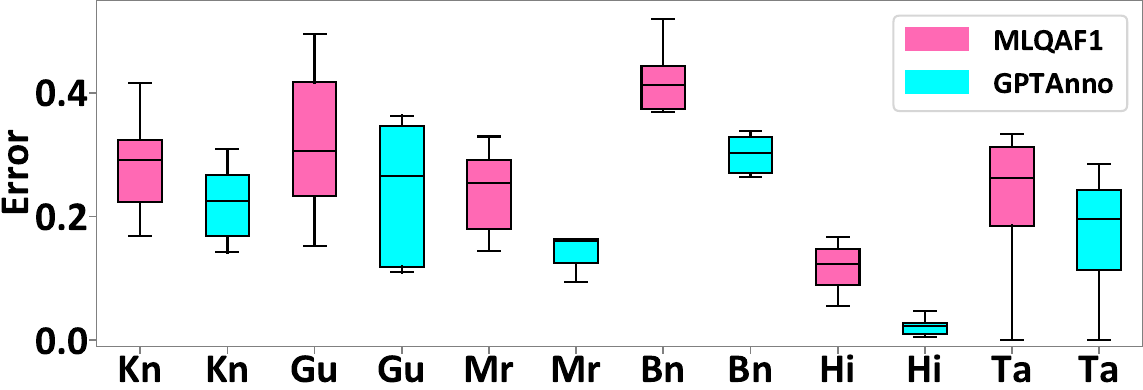}
        \caption{Comparison of \meta and \gptevalf. }
        \label{fig:error_metrics_ha}
        \vspace{-12pt}
\end{wrapfigure}
We validated GPTAnnotator- F1 by selecting 100 questions from the \indic dataset (across six languages) and comparing the results with human annotations (\haf). 	Human annotators were native speakers and were given clear instructions for annotations. 
As shown in Figure~\ref{fig:error_metrics_ha} \meta scores differed from human annotations by an average of 25\% (with a maximum of 51\%), exposing the limitations of current GTs. In contrast, \gptevalf reduced the error difference by 30\%, aligning more closely to human judgment. Thus, providing a more accurate reflection of LLM performance. In the subsequent sections, we present results using both \meta and \gptevalf metrics.

\textbf{Limitations of \gpteval: } GPTAnnotator's quality depends on the LLM's performance in the target language. Designers should check benchmarks/baselines or run small-scale human evaluations to compare annotations. If discrepancies are significant, the LLM may not be suitable. Despite some limitations, our evaluations show GPTAnnotator aligns well with human annotators, proving effective across languages.

\begin{table*}[]
\centering
\resizebox{0.85\textwidth}{!}{%
\begin{tabular}{llll|lll}
\hline
 & \multicolumn{3}{c|}{\meta}                          & \multicolumn{2}{c}{\gptevalf} &                     \\ \hline
 & \gptt & \turbo & \mix & \gptt & \turbo & \mix \\ \hline
\mono     & \textbf{0.51} & \textbf{0.43} & 0.15 & 0.71 & 0.71 & 0.31 \\
\trans     & 0.36 & 0.37 & \textbf{0.33} & \textbf{0.80} & \textbf{0.80} & \textbf{0.68} \\
\siml     & 0.30 & 0.28 & 0.19 & 0.70 & 0.70 & 0.44 \\
\aggsrc   & \textbf{0.51} & \textbf{0.43} & 0.20 & 0.73 & 0.73 & 0.39 \\
\aggtrans & 0.35 & 0.38 & \textbf{0.33} & 0.79 & 0.79 & \textbf{0.68} \\ \hline
\end{tabular}%
}
\caption{Performance of different Prompting strategies for \indic. }
\label{tab:prompt_indic}
\vspace{-10pt}
\end{table*}
\section{Prompt Strategies for Polyglot LLMs}
\label{sec:prompts}
Effective prompt design is critical for improving generative models, especially in multilingual tasks~\cite{sahoo2024systematic}. Crafting prompts is already challenging in monolingual English~\cite{yang2022prompt}, and becomes more complex across diverse languages due to differences in syntax, grammar, and lexicon. Various prompt strategies have been proposed, each with its advantages and limitations across languages~\cite{mega,ahuja2023megaverse}.

Chain-of-Thought prompting~\cite{weichain,lai2024mcot} excels in reasoning tasks but struggles with morphologically complex languages like Korean. Self-Translation~\cite{gao2024llms}, where models refine responses across languages, can cause inconsistencies, particularly in low-resource languages. Linguistic Feature Prompting~\cite{nie2024decomposed,messina2023prompting} encodes syntactic or semantic features directly into prompts, aiding models in languages with complex grammar. Finally, Aggregation strategies, which combine responses from multiple prompt types, offer a way to mitigate prompt-specific weaknesses~\cite{wang2023multiple,lin2021few}.

However, no single strategy works best across all languages. Success depends on factors such as language-specific traits, task complexity, and resource availability.

\textbf{Selected Strategies from Prior Work.}
From the variety of prompt strategies available, we selected five that consistently showed the best performance across tasks and languages. These strategies may not be universally optimal, but they provide strong results in our multilingual experiments.

\begin{table*}[!t]
\begin{minipage}[l]{0.33\linewidth}
\renewcommand{\arraystretch}{3.0}
\resizebox{\columnwidth}{!}{%
\begin{tabular}{llllll}
\hline
Metrics                                   & Models                & Ada  & Adav3 & XLMR & Cohere \\ \hline
\multirow{2}{*}{\meta}     & GPT4T  & 0.51 & 0.5   & 0.54 & \textbf{0.58}   \\
                                          & GPT3.5 & 0.43 & 0.43  & 0.39 & \textbf{0.44}   \\ \hline
\multirow{2}{*}{GPTAnno} & GPT4T  & 0.8  & 0.8   & 0.8  & \textbf{0.82}   \\
                                          & GPT3.5 & 0.8  & 0.8   & 0.8  & \textbf{0.81}   \\ \hline
\end{tabular}%
}
\caption{Hybrid approach performance on \indic.}
\label{tab:hybrid-perf}
        \vspace{-5pt}
\end{minipage}
~~
\begin{minipage}[l]{0.3\linewidth}
\renewcommand{\arraystretch}{1.5}
\resizebox{0.9\textwidth}{!}{%
\begin{tabular}{l|ll|ll}
\hline
     & \multicolumn{2}{l|}{\meta}    & \multicolumn{2}{l}{\gptevalf} \\ \hline
Lang & GPT3.5 & GPT4T & GPT3.5 & GPT4T  \\ \hline
as & Ada    & Cohere & Ada    & Ada    \\
bn & Cohere & Cohere & Ada    & Ada    \\
gu & Ada    & Cohere & Cohere & Cohere \\
hi & Cohere & Cohere & Ada    & Cohere \\
kn & Ada    & Cohere & Cohere & Cohere \\
ml & Cohere & Cohere & Cohere & Cohere \\
mr & Ada    & Cohere & Cohere & Cohere \\
or & Adav3   & Ada    & Ada    & Ada    \\
pa & Adav3   & Adav3   & Cohere & Ada    \\
ta & Cohere & Cohere & Cohere & Cohere \\
te & Cohere & Cohere & Cohere & Cohere \\ \hline
\end{tabular}%
}
\caption{Embedding preference \indic.}
\label{tab:emd-pre-indic}
        \vspace{-5pt}
\end{minipage}
~~
\begin{minipage}[l]{0.31\linewidth}
\renewcommand{\arraystretch}{1.5}
\resizebox{1.0\textwidth}{!}{%
\begin{tabular}{l|ll|ll}
\hline
 & \multicolumn{2}{l|}{\meta}    & \multicolumn{2}{l}{\gptevalf} \\ \hline
Lang & GPT3.5 & GPT4T & GPT3.5 & GPT4T \\ \hline
ar & Ada    & Adav3    & Ada    & Adav3    \\
bn & Ada    & Ada     & Ada    & Ada     \\
en & Cohere & XLMR & Cohere & XLMR \\
fi & Ada    & Ada     & Ada    & Ada     \\
id & Ada    & Adav3    & Ada    & Adav3    \\
ko & Ada    & XLMR & Ada    & XLMR \\
ru & Ada    & Ada     & Ada    & Ada     \\
sw & Ada    & Adav3    & Ada    & Adav3    \\
te & Ada    & Cohere  & Ada    & Cohere \\ \hline
\end{tabular}%
}
\caption{Embedding preference \tydi.}
\label{tab:emd-pre-tydi}
        \vspace{-5pt}
\end{minipage}
\vspace{-5pt}
        \vspace{-10pt}
\end{table*}

\textbf{1. Monolingual (Mono):} Instruction, context, and examples are provided in the source language. This works well for high-resource languages but underperforms for low-resource ones~\cite{mega}.

\textbf{2. Translate-Test (Trans):} Instructions and contexts are translated into English, leveraging the model’s strengths in English before back-translating the output to the source language. However, translation errors can affect accuracy in low-resource languages~\cite{agrawal2024translation,ghafoor2021impact}.

\textbf{3. Similar High-Resourced Language (Sim): }Roundtripping through a linguistically similar high-resource or medium-resource language (chosen based on lang2vec~\cite{littell2017uriel}) improves performance by capturing linguistic similarities better than direct English translation, especially for related languages. More details in Appendix~\ref{app:sim}.

\textbf{4. Aggregation Source (Agg\_Src):} Combines responses from multiple strategies (Mono, Trans, Sim) to form a final answer in the source language. Though computationally expensive, this leads to more coherent, accurate answers~\cite{wang2023multiple}.

\textbf{5. Aggregation Translate (Agg\_Trans):} Aggregates responses in English before back-translating to the source language. While translation challenges exist, high-quality translation systems make this approach effective.

Other approaches like self-translation and linguistic feature-based prompting are viable but didn’t perform consistently. We tested both zero-shot and few-shot setups, finding that few-shot examples consistently improved performance. Future advancements in example selection and in-context learning will further enhance these strategies.

\textbf{Prompting Strategies Results.}
Our results highlight three key findings:

\textbf{1. No Universal Best Strategy:} No single prompt strategy works best across all models and languages. For \gptt and \turbo, \mono and \aggsrc excel, while \mix favors \trans and \aggtrans. Translation-based strategies work better for low-resource languages like Tamil and Telugu due to limited data availability. \\
\textbf{2. Strategy Sensitivity to Metrics:} Performance varies based on the evaluation metric. For example, \gptevalf favors \trans and \aggtrans, while \meta shows better results with \mono and \aggsrc. \\
\textbf{3. Comparable Performance with Metric Variation:} While \meta suggests \gptt outperforms \turbo, enriching ground truth and using \gptevalf reveals comparable performance, with a 28\% overall improvement in accuracy for \turbo. This underscores the importance of metric selection when evaluating models. \\
\textbf{Summary:} \textit{Prompt strategies significantly boost multilingual model performance, but no single approach is universally superior across models, metrics, or languages. The \gptevalf metric, in particular, levels the performance gap between \turbo and \gptt.}

\section{Hybrid Approach: Synthesizing LLM Generation with Multilingual Embeddings}
\label{sec:hybrid}

While LLMs excel in response synthesis, improving multilingual performance requires robust multilingual embeddings. GPT models, primarily trained on English data, use the default embedding model (text-embedding-ada-002, or ada), which underperforms in multilingual contexts. In contrast, state-of-the-art multilingual models like XLMR-XXL~\cite{goyal2021larger} and Cohere (embed-multilingual-v3.0)~\cite{cohere2024embedv3} demonstrate superior results due to their diverse language training.

We leverage a hybrid approach that combines the cross-lingual understanding of multilingual embeddings with the text-generation abilities of LLMs. We experiment with GPT's default ada embeddings, an improved variant (adav3)~\cite{openai2024ada3}, and state-of-the-art multilingual embeddings like XLMR-XXL~\cite{goyal2021larger} and Cohere v3~\cite{cohere2024embedv3}.

\textbf{Performance Analysis.} Table~\ref{tab:hybrid-perf} illustrates the maximum performance achieved by each embedding (ada, adav3, xlmr, cohere) for \gptt and \turbo models across all languages and prompt strategies for \indic. Cohere, a multilingual embedding, enhances \gptt performance by up to 7\% and 2\% compared to default ada embeddings when using \meta and \gptevalf metrics. This indicates a significant improvement in multilingual task performance with multilingual embeddings coupled with LLM generation. While marginal improvements are observed in \turbo with multilingual embeddings, mainly due to poor LLM generation with \turbo rather than multilingual content retrieval. 

Additionally, Table~\ref{tab:emd-pre-indic} and~\ref{tab:emd-pre-tydi} indicates the preferred embedding for each language that yields the best performance. Generally, multilingual embeddings, particularly Cohere, are preferred for \indic. Similar trends are observed in \tydi, as detailed in Appendix~\ref{app:hybrid}. 
 
\textbf{Summary:} \textit{The hybrid approach boosts performance by up to 7\% on the \gptt model. However, there's no universal best prompt strategy, model, or embedding that performs optimally across datasets and languages.}

\begin{figure*}[t!]\centering
    \includegraphics[width=0.759\linewidth]{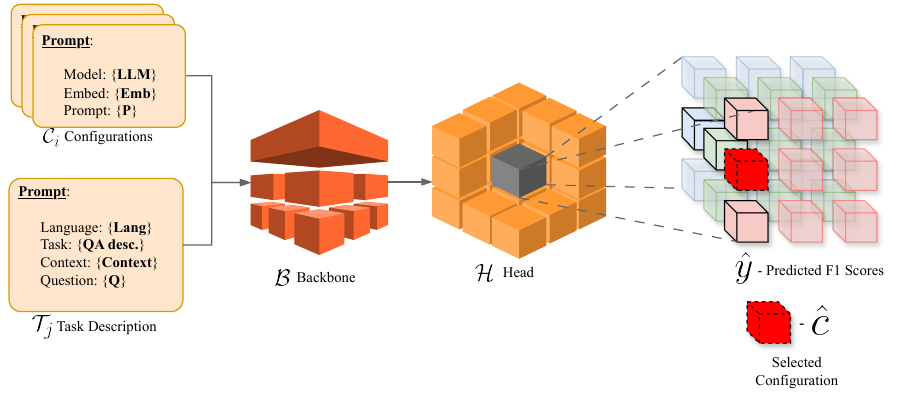}
    \vspace{-15pt}
    \caption{Illustration of Inference Pipeline}
    \label{app:fig:inference}
\end{figure*}
\section{Dynamic Learning Approach to Improve Multilingual Performance}
\label{sec:learning}
\vspace{-5pt}
\textbf{Motivation:} A one-size-fits-all solution does not exist for selecting the best combination of prompt strategy, embeddings, and LLM for different languages. This raises the key question: Can we dynamically determine the optimal configuration for each query to maximize multilingual performance?

To address this, we propose a learning approach that dynamically selects the optimal configuration per query, meeting three key requirements: (i) Offline Learning: It learns the best configuration using ground truth data offline, (ii) Online Learning: It adapts in real-time, adjusting for new data and distribution shifts, and (iii) Language and Dataset Adaptability: It remains flexible across languages and datasets, ensuring robust performance.

\textbf{Hybrid Architecture:} Our solution combines LLMs with convolutional layers to dynamically select the best configuration across LLM models, embeddings, and prompt strategies. LLMs generate high-dimensional representations, which are fed into ND convolutional layers that extract features across dimensions, predicting task accuracy (F1 score) per query. By comparing predicted scores, we select the optimal configuration for each task and language, for both offline and online learning.

Prior efforts like LOVM~\cite{zohar2024lovm}, \cite{liu2023towards} and HuggingGPT~\cite{shen2024hugginggpt} focus on optimizing model selection for a single parameter. In contrast, our approach selects the optimal combination of LLM model, prompt strategy, and embeddings, tackling a complex, high-dimensional search space.

In our architecture, we predict F1 scores for each configuration, generating a SoftMax output as a probability distribution. Sampling configurations from this distribution in online settings allows for controlled entropy and exploration of diverse configurations, helping mitigate bias, especially with out-of-distribution data.

\begin{table*}[!t]
\begin{minipage}[l]{0.5\linewidth}
\renewcommand{\arraystretch}{2}
\resizebox{\textwidth}{!}{%
\begin{tabular}{|l|l|cc|cc|ccc|}
\hline
Evaluation & Datasets             & Acc & Acc & F1& F1       & Max           & Random & Best \\ 
& &@top1 & @top5& @top1&@top5 & F1& F1& single F1\\\hline
\multirow{2}{*}{\meta}     & \indic & 0.41 & 0.83 & 0.60 & \textbf{0.64} & 0.64          & 0.46 & 0.51 \\
           & \tydi & 0.57     & 0.78     & 0.52    & \textbf{0.54} & 0.54          & 0.43   & 0.50        \\ \hline
\multirow{2}{*}{GPTAnno} & \indic & 0.32 & 0.48 & 0.59 & 0.68          & \textbf{0.69} & 0.49 & 0.58 \\
           & \tydi & 0.62     & 0.55     & 0.56    & 0.69          & \textbf{0.72} & 0.51   & 0.54        \\ \hline
\end{tabular}%
}
\caption{Offline performance.}
\label{tab:offline}
        \vspace{-5pt}
\end{minipage}
\begin{minipage}[l]{0.5\linewidth}
\renewcommand{\arraystretch}{2}
\resizebox{\textwidth}{!}{%
\begin{tabular}{|l|l|cc|cc|ccc|}
\hline
Evaluation                            & Datasets              & Acc   & Acc   & F1    & F1            & Max           & Random & Best   \\
                                      &                       & @top1 & @top5 & @top1 & @top5         &  F1             &   F1     & single F1 \\ \hline
\multirow{2}{*}{\meta} & \indic & 0.29  & 0.73  & 0.60  & \textbf{0.63} & 0.63          & 0.46   & 0.51   \\
                                      & \tydi  & 0.62  & 0.85  & 0.51  & \textbf{0.52} & 0.52          & 0.41   & 0.45   \\ \hline
\multirow{2}{*}{GPTAnno}              & \indic & 0.52  & 0.57  & 0.62  & 0.66          & \textbf{0.69} & 0.52   & 0.61   \\
                                      & \tydi  & 0.62  & 0.67  & 0.73  & 0.74          & \textbf{0.76} & 0.54   & 0.69   \\ \hline
\end{tabular}
}
\caption{Online performance.}
\label{tab:online}
        \vspace{-5pt}
\end{minipage}
\vspace{-5pt}
        \vspace{-10pt}
\end{table*}

\textbf{Architecture details.} The architecture leverages the LLaMa-2-70B-hf model for embedding generation. The traditional sampling head is replaced by a set of Conv-ND layers \cite{vizcaino2021convnd}, denoted as $\mathcal{H}$, which predict the F1 Score for each configuration. The LLaMa-2-70B-hf~\cite{touvron2023llama} backbone, $\mathcal{B}$, embeds the Task Description $\mathcal{T}$, into task embeddings $\mathcal{E}_T$and configuration embeddings $\mathcal{C}_i$ into $\mathcal{E}_{C_i}$.

These embeddings are then arranged into an ND array of size $\mathcal{R}^{e \times n_1 \times n_2 \times n_3 \dots n_m}$, where $m$ is the number of parameters (e.g., language model, embedding model, prompt strategies, so $m=3$).
Each $n_i$ represents the number of possibilities for each parameter (e.g., three LLMs (\gptt, \turbo, \mix), four embedding models (adav2,adav3,XLMR,cohere), five prompt strategies (\mono, \trans, \siml, \aggsrc, \aggtrans)). The embedding projection size $e$ for $\mathcal{B}$ is 8192. The task embedding is broadcasted and concatenated with configuration embeddings to form a matrix of size $\mathcal{R}^{2e \times n_1 \times n_2 \times n_3 \dots n_m}$. 

We treat the embedding dimension as the number of input channels to $\mathcal{H}$ and reduce it to 1 while preserving the remaining dimensions, resulting in a matrix of size $\mathcal{R}^{1 \times n_1 \times n_2 \times n_3 \dots n_m}$ or $\mathcal{R}^{n_1 \times n_2 \times n_3 \dots n_m}$, representing the predicted F1 for all configurations.

\begin{gather}
\mathcal{E}_{T_j} \leftarrow \mathcal{B}(\mathcal{T}_j)\\
\mathcal{E}_{C_i} \leftarrow \mathcal{B}(\mathcal{C}_i)\\
\mathcal{E}_j \leftarrow \mathcal{E}_{T_j} \parallel \mathcal{E}_{C_i} \\
\hat{y} \leftarrow \mathcal{H}(\mathcal{E}_j)
\end{gather}
Using the above equations, we obtain \(\hat{y}\), which is the predicted F1 score for all combinations. To select the configuration, we either take the \emph{argmax} or apply \emph{softmax} and sample a particular configuration. Figure~\ref{app:fig:inference} illustrates inference pipeline, given the Task Description \(\mathcal{T}_{j}\) and Configurations \(\mathcal{C}_{i}\) to obtain \(\hat{c}\) for sampled configuration.

\subsection{Training the Model for Both Online and Offline Setups.} We train the Backbone \(\mathcal{B}\) and the Head \(\mathcal{H}\) using different loss functions for offline and online settings.

\textbf{1. Offline Setting:} In the offline setting, we have the advantage of knowing the F1 scores for all possible configurations for each given sample. This complete information allows us to obtain the ground truth F1 scores for all samples, denoted as \(y\). We can then use these ground truth F1 scores to train the backbone \(\mathcal{B}\) and the head \(\mathcal{H}\) effectively.

(a) \textbf{Infer F1 Scores for All Configurations}: For each sample, infer the F1 scores for all possible configurations. For example, if there are three configurations for each parameter (e.g., three language models (\gptt, \turbo, \mix), four embedding models (adav2,adav3,XLMR,cohere), five prompt strategies (\mono,\trans, \siml,\aggsrc, \aggtrans)), we would infer F1 scores for \(3 \times 4 \times 5 = 60\) configurations per sample.

(b) \textbf{Obtain Ground Truth F1 Scores}: Collect the actual F1 scores for all configurations, which serve as the ground truth \(y\). Thus, for each sample, we gather the F1 scores for all 60 configurations.

(c) \textbf{Train Using MSE Loss}: Use the Mean Squared Error (MSE) loss to train the model. The MSE loss is computed between the predicted F1 scores \(\hat{y}\) and the ground truth F1 scores \(y\)  $   {MSE Loss} = \frac{1}{N} \sum_{i=1}^N (\hat{y}_i - y_i)^2     $,    where \(N\) is the number of samples.

\textbf{2. Online Setting:} In the online setting, we only have the ground truth F1 score for the configuration that was selected and inferred. This results in a sparse matrix of F1 scores, as we do not compute the F1 scores for all configurations to avoid the computational cost. 

(a) \textbf{Infer F1 Score for Selected Configuration}: For each sample, infer the F1 score for only the selected configuration. This selected configuration is chosen based on the model's predictions or a sampling strategy. For example, if the model predicts a specific configuration out of 60, we only compute the F1 score for that particular configuration. 

(b) \textbf{Obtain Ground Truth F1 Score}: Compute the actual F1 score for the selected configuration, which serves as the ground truth \(y_{{selected}}\). 

(c) \textbf{Update Using Sparse Matrix}: Update the model using the sparse matrix of predicted F1 scores \(\hat{y}\). Only the score corresponding to that configuration is updated, leaving other entries unaffected, thus reducing computational overhead.

(d) \textbf{Adjust Loss Function}: The loss function must account for the sparsity. Instead of a straightforward MSE loss, we use a modified loss function that updates only the predicted F1 score for the selected configuration, 
 ${Sparse MSE Loss} = (\hat{y}_{{selected}} - y_{{selected}})^2 $, where \(\hat{y}_{{selected}}\) is the predicted F1 score for the selected configuration, and \(y_{{selected}}\) is the ground truth F1 score for the same configuration. This approach optimizes the model without needing to compute F1 scores for all configurations, reducing computational overhead. Implementation details of the above pipeline is explained in Appendix~\ref{app:training}.

\textbf{Train-Test Split.} The datasets are divided into three parts: 60\% for offline training, 20\% for online adaptation, and 20\% for testing. Evaluation is performed using the \meta and \gptevalf metrics.

In our online setup evaluation, we use ground truth due to the difficulty of collecting real-time user feedback, however this setup mirrors an online active learning environment where feedback is gathered on specific samples. As this solution is deployed in QA chatbots or copilots, user feedback would allow it to adapt to new scenarios over time.

\subsection{Evaluation of Learning Approach}
\textbf{1. Offline Training Results.}
We evaluated our model against two baselines: (i) Random Configuration Selection, and (ii) Best Single Configuration (the highest-scoring configuration for all samples). Performance was measured using Accuracy (Acc@Top1, Acc@Top5) and F1 score at top 1 and top 5 configurations.


As shown in Table~\ref{tab:offline} our model outperforms random selection by 17\% and the best single configuration by 11\%, consistently across both \meta and \gptevalf scores. Notably, our approach achieves a top 5 accuracy that matches the maximum achievable accuracy, underscoring its robustness in generating correct answers.

\textbf{2. Online Training results:} We evaluated our model’s adaptability to new data distributions by further training it for 10 epochs using parameters from the offline phase (epoch 100). In online adaptation, our model achieved top 1 and top 5 F1 scores of 60\% and 63\%, respectively, closely approaching the maximum accuracy (63\%) (see Table \ref{tab:online}). It outperformed random selection by 15\% and the best single configuration by 7\%, demonstrating effectiveness even with minimal fine-tuning on new or out-of-distribution data.


\begin{table}[]
\centering
\renewcommand{\arraystretch}{2.5}
\resizebox{0.5\textwidth}{!}{%
\begin{tabular}{l|l|ll|ll|lll}
\hline
Evalaution & Languages & Acc@top1 & Acc@top5 & F1@top1 & F1@top5       & Max-
F1           & Random-F1 & Best Single-F1 \\ \hline
\multirow{3}{*}{\begin{tabular}[c]{@{}l@{}}Language\\ Adaptation\end{tabular}} &
  Kn &
  0.29 &
  0.75 &
  0.44 &
 \textbf{{0.46}} &
  {0.47} &
  0.37 &
  0.45 \\
           & Ta        & 0.28     & 0.74     & 0.48    & \textbf{{0.50}} & {0.53} & 0.43   & 0.46        \\
           & Te        & 0.28     & 0.74     & 0.51    & \textbf{{0.55}} & {0.57} & 0.43   & 0.49        \\ \hline
\begin{tabular}[c]{@{}l@{}}Dataset \\ Adaptation\end{tabular} &
  \begin{tabular}[c]{@{}l@{}}\tydi on  \\ \indic base\end{tabular} &
  0.56 &
  0.67 &
  0.43 &
  \textbf{{0.52}} &
  {0.52} &
  0.41 &
  0.45 \\ \hline
\end{tabular}%
}
\caption{Learning approach performance on adaptation to unseen languages and datasets.}
\label{tab:adapt}
\vspace{-10pt}
\end{table}

\textbf{3. Adaptation Efficacy:} 
\textit{(i) Adaptation to Unseen Languages:} We tested the model’s ability to adapt to languages not encountered during offline training. We trained the backbone and the head on the \indic dataset, excluding Kannada, Tamil, and Telugu languages. The excluded languages were then used for online training, simulating scenarios where the model encounters new languages during inference. 
Results in Table~\ref{tab:adapt} show the model generalizing effectively, achieving F1 scores close to the maximum, and outperforming baselines across all languages, proving its adaptability in multilingual scenarios.
\textit{(ii) Adaptation to Different Datasets:} We also assessed adaptation to different datasets by training on the \indic dataset and testing on \tydi. Despite limited language overlap, our model exceeded random selection by 11\% and the best single configuration by 7\%. With just 15 fine-tuning epochs on 20\% of the \tydi dataset, it achieved the maximum F1 score, reinforcing its ability to handle diverse datasets and query distributions.

\textbf{Summary:} \textit{Our approach demonstrates substantial improvements in dynamically selecting configurations and adapting to new languages and datasets, showcasing its effectiveness and adaptability in real-world multilingual applications.}

\subsection{Comparing with Language specific fine-tuned model}
We conducted extensive experiments comparing our dynamic learning approach with state-of-the-art (SOTA) fine-tuned language-specific models. Remarkably, our approach outperforms these fine-tuned models by over 4x in terms of F1 scores.

For example, the Navarasa and Aryabhatta models, fine-tuned on over 10 Indian languages, achieve an average F1 score of just 10\% on the Indic dataset across all languages. In contrast, our dynamic approach, as shown in Tables~\ref{tab:offline} and~\ref{tab:online}, consistently achieves 60-70\% F1 scores. Similarly, we evaluated bi-lingual models like Ambari (fine-tuned for Kannada) and Airavata (fine-tuned for Hindi) on the Indic QA dataset, where their F1 scores were below 5\%. This highlights the limitations of fine-tuned models in handling real-world QA tasks.

In contrast, our dynamic approach, without language-specific fine-tuning, achieves F1 scores of over 50-60\% across various languages, even when the model was not trained on those specific languages. For instance, as shown in Table~\ref{tab:adapt}, our model achieved an F1 score of 46\% on Kannada (KA) without prior training, while Ambari scored less than 2\%. This demonstrates the broad applicability and superior performance of our approach across diverse languages, applications, and tasks.

\textbf{Practical usage: } To use our dynamic algorithm, users provide three inputs: (a) base LLM models, (b) multilingual embeddings, and (c) prompt strategies. Our system then dynamically selects the best combination of these for the given language and task, optimizing multilingual performance without manual fine-tuning.
\section{Conclusions}

In this work, we introduced a dynamic learning framework to enhance multilingual LLM performance without extensive training or fine-tuning. Our findings show that prompting strategies are not universally effective, requiring dynamic, language-specific approaches to optimize performance across datasets, models, and languages. Second, our hybrid use of multilingual embeddings, particularly with Cohere, achieved up to a 7\% performance boost on the \gptt model, highlighting the importance of embedding selection in cross-lingual understanding. Most notably, our dynamic runtime configuration framework demonstrated 10-15\% improvements in multilingual task performance and up to 4x gains over language-specific fine-tuned models. Our framework outperformed static, best configurations and baseline models, proving its effectiveness in both offline and online settings. This dynamic adaptability not only enhances LLMs' multilingual capabilities but also future-proofs them, allowing seamless integration with emerging models and strategies. Future research directions include exploration of learning techniques, scalability to larger datasets, and the generalization of our approach to other tasks. 


\textbf{Limitations and Broader Research:} 
While our work takes a first step towards improving multilingual performance, the system is still not fully inclusive, and as a community, we must explore ways to ensure LLMs are accessible to all. Finally, while our key contributions including learning algorithms are generalizable, the optimal strategies and embeddings may differ from one dataset to another. With the growing demand for multilingual language models, our findings pave the way for future advancements in Polyglot LLM performance.
\bibliography{ref}

\begin{thebibliography}{60}
\providecommand{\natexlab}[1]{#1}

\bibitem[{Abdin et~al.(2024)Abdin, Jacobs, Awan, Aneja, Awadallah, Awadalla, Bach, Bahree, Bakhtiari, Behl et~al.}]{abdin2024phi}
Marah Abdin, Sam~Ade Jacobs, Ammar~Ahmad Awan, Jyoti Aneja, Ahmed Awadallah, Hany Awadalla, Nguyen Bach, Amit Bahree, Arash Bakhtiari, Harkirat Behl, et~al. 2024.
\newblock Phi-3 technical report: A highly capable language model locally on your phone.
\newblock \emph{arXiv preprint arXiv:2404.14219}.

\bibitem[{Agrawal et~al.(2024)Agrawal, Fazili, and Jyothi}]{agrawal2024translation}
Ashish~Sunil Agrawal, Barah Fazili, and Preethi Jyothi. 2024.
\newblock Translation errors significantly impact low-resource languages in cross-lingual learning.
\newblock \emph{arXiv preprint arXiv:2402.02080}.

\bibitem[{Ahuja et~al.(2023{\natexlab{a}})Ahuja, Hada, Ochieng, Jain, Diddee, Maina, Ganu, Segal, Axmed, Bali et~al.}]{mega}
Kabir Ahuja, Rishav Hada, Millicent Ochieng, Prachi Jain, Harshita Diddee, Samuel Maina, Tanuja Ganu, Sameer Segal, Maxamed Axmed, Kalika Bali, et~al. 2023{\natexlab{a}}.
\newblock Mega: Multilingual evaluation of generative ai.
\newblock \emph{arXiv preprint arXiv:2303.12528}.

\bibitem[{Ahuja et~al.(2023{\natexlab{b}})Ahuja, Aggarwal, Gumma, Watts, Sathe, Ochieng, Hada, Jain, Axmed, Bali et~al.}]{ahuja2023megaverse}
Sanchit Ahuja, Divyanshu Aggarwal, Varun Gumma, Ishaan Watts, Ashutosh Sathe, Millicent Ochieng, Rishav Hada, Prachi Jain, Maxamed Axmed, Kalika Bali, et~al. 2023{\natexlab{b}}.
\newblock Megaverse: benchmarking large language models across languages, modalities, models and tasks.
\newblock \emph{arXiv preprint arXiv:2311.07463}.

\bibitem[{AI(2023)}]{anthropic2023claude3}
Anthropic AI. 2023.
\newblock \href {https://www-cdn.anthropic.com/de8ba9b01c9ab7cbabf5c33b80b7bbc618857627/Model_Card_Claude_3.pdf} {Model card for claude 3}.
\newblock Accessed: 2024-05-21.

\bibitem[{AI(2024)}]{sarvamai2024openhathi}
Sarvam AI. 2024.
\newblock \href {https://www.sarvam.ai/blog/announcing-openhathi-series} {Announcing openhathi series}.
\newblock Sarvam AI Blog.
\newblock Accessed: 2024-09-15.

\bibitem[{AI4Bharat(2022)}]{ai4bharat2022indicqa}
AI4Bharat. 2022.
\newblock Indicqa: A multilingual question answering dataset for 12 indic languages.
\newblock \url{https://huggingface.co/datasets/ai4bharat/IndicQA}.

\bibitem[{Brown et~al.(2020)Brown, Mann, Ryder, Subbiah, Kaplan, Dhariwal, Neelakantan, Shyam, Sastry, Askell, Agarwal, Herbert{-}Voss, Krueger, Henighan, Child, Ramesh, Ziegler, Wu, Winter, Hesse, Chen, Sigler, Litwin, Gray, Chess, Clark, Berner, McCandlish, Radford, Sutskever, and Amodei}]{DBLP:journals/corr/abs-2005-14165}
Tom~B. Brown, Benjamin Mann, Nick Ryder, Melanie Subbiah, Jared Kaplan, Prafulla Dhariwal, Arvind Neelakantan, Pranav Shyam, Girish Sastry, Amanda Askell, Sandhini Agarwal, Ariel Herbert{-}Voss, Gretchen Krueger, Tom Henighan, Rewon Child, Aditya Ramesh, Daniel~M. Ziegler, Jeffrey Wu, Clemens Winter, Christopher Hesse, Mark Chen, Eric Sigler, Mateusz Litwin, Scott Gray, Benjamin Chess, Jack Clark, Christopher Berner, Sam McCandlish, Alec Radford, Ilya Sutskever, and Dario Amodei. 2020.
\newblock \href {https://arxiv.org/abs/2005.14165} {Language models are few-shot learners}.
\newblock \emph{CoRR}, abs/2005.14165.

\bibitem[{Clark et~al.(2020)Clark, Choi, Collins, Garrette, Kwiatkowski, Nikolaev, and Palomaki}]{tydiqa}
Jonathan~H. Clark, Eunsol Choi, Michael Collins, Dan Garrette, Tom Kwiatkowski, Vitaly Nikolaev, and Jennimaria Palomaki. 2020.
\newblock Tydi qa: A benchmark for information-seeking question answering in typologically diverse languages.
\newblock \emph{Transactions of the Association for Computational Linguistics}.

\bibitem[{Cohere(2024)}]{cohere2024embedv3}
Cohere. 2024.
\newblock \href {https://cohere.com/blog/introducing-embed-v3} {Introducing embed v3}.
\newblock Cohere Blog.
\newblock Accessed: 2024-05-21.

\bibitem[{corporation()}]{azureopenai}
Microsoft corporation.
\newblock Azure openai service.
\newblock \url{https://azure.microsoft.com/en-us/products/cognitive-services/openai-service}.

\bibitem[{FarmerChat(2024)}]{farmerchat}
Digital~Green FarmerChat. 2024.
\newblock \href {https://farmerchat.digitalgreen.org/} {Farmer chat}.
\newblock Website.
\newblock Accessed: 2024-05-21.

\bibitem[{Fu et~al.(2022)Fu, Ng, and Liu}]{fu2022polyglot}
Jinlan Fu, See-Kiong Ng, and Pengfei Liu. 2022.
\newblock Polyglot prompt: Multilingual multitask promptraining.
\newblock \emph{arXiv preprint arXiv:2204.14264}.

\bibitem[{Gala et~al.(2024)Gala, Jayakumar, Husain, Khan, Kanojia, Puduppully, Khapra, Dabre, Murthy, Kunchukuttan et~al.}]{gala2024airavata}
Jay Gala, Thanmay Jayakumar, Jaavid~Aktar Husain, Mohammed Safi Ur~Rahman Khan, Diptesh Kanojia, Ratish Puduppully, Mitesh~M Khapra, Raj Dabre, Rudra Murthy, Anoop Kunchukuttan, et~al. 2024.
\newblock Airavata: Introducing hindi instruction-tuned llm.
\newblock \emph{arXiv preprint arXiv:2401.15006}.

\bibitem[{Gao et~al.(2024)Gao, Chen, Chen, Dai, Jin, Jiang, Ning, Yu, Xuan, Cai et~al.}]{gao2024llms}
Dehong Gao, Kaidi Chen, Ben Chen, Huangyu Dai, Linbo Jin, Wen Jiang, Wei Ning, Shanqing Yu, Qi~Xuan, Xiaoyan Cai, et~al. 2024.
\newblock Llms-based machine translation for e-commerce.
\newblock \emph{Expert Systems with Applications}, 258:125087.

\bibitem[{Ghafoor et~al.(2021)Ghafoor, Imran, Daudpota, Kastrati, Batra, Wani et~al.}]{ghafoor2021impact}
Abdul Ghafoor, Ali~Shariq Imran, Sher~Muhammad Daudpota, Zenun Kastrati, Rakhi Batra, Mudasir~Ahmad Wani, et~al. 2021.
\newblock The impact of translating resource-rich datasets to low-resource languages through multi-lingual text processing.
\newblock \emph{IEEE Access}, 9:124478--124490.

\bibitem[{Goyal et~al.(2021)Goyal, Du, Ott, Anantharaman, and Conneau}]{goyal2021larger}
Naman Goyal, Jingfei Du, Myle Ott, Giri Anantharaman, and Alexis Conneau. 2021.
\newblock Larger-scale transformers for multilingual masked language modeling.
\newblock \emph{arXiv preprint arXiv:2105.00572}.

\bibitem[{Huang et~al.(2024)Huang, Mo, Li, Li, Zhang, Yi, Mao, Liu, Xu, Xu et~al.}]{huang2024survey}
Kaiyu Huang, Fengran Mo, Hongliang Li, You Li, Yuanchi Zhang, Weijian Yi, Yulong Mao, Jinchen Liu, Yuzhuang Xu, Jinan Xu, et~al. 2024.
\newblock A survey on large language models with multilingualism: Recent advances and new frontiers.
\newblock \emph{arXiv preprint arXiv:2405.10936}.

\bibitem[{Huang et~al.(2022)Huang, Ma, Zhang, Wei, and Wang}]{huang2022zero}
Lianzhe Huang, Shuming Ma, Dongdong Zhang, Furu Wei, and Houfeng Wang. 2022.
\newblock Zero-shot cross-lingual transfer of prompt-based tuning with a unified multilingual prompt.
\newblock \emph{arXiv preprint arXiv:2202.11451}.

\bibitem[{HuggingFace({\natexlab{a}})}]{Airavata}
HuggingFace. {\natexlab{a}}.
\newblock Airavata: Introducing hindi instruction-tuned llm.
\newblock \url{https://huggingface.co/ai4bharat/Airavata}.

\bibitem[{HuggingFace({\natexlab{b}})}]{Ambari}
HuggingFace. {\natexlab{b}}.
\newblock Ambari: A series of open source bilingual kannada-english large language models.
\newblock \url{https://huggingface.co/collections/Cognitive-Lab/ambari-65a2678d1051c2b0db3e01fe}.

\bibitem[{HuggingFace({\natexlab{c}})}]{Navarasa}
HuggingFace. {\natexlab{c}}.
\newblock Navarasa 2.0 models: Collection of models navarasa 2.0 models finetuned with gemma on 15 indian languages.
\newblock \url{https://huggingface.co/collections/Telugu-LLM-Labs/navarasa-20-models-65f7c72addf0619cb0991309}.

\bibitem[{Hämmerl et~al.(2022)Hämmerl, Deiseroth, Schramowski, Libovický, Fraser, and Kersting}]{hämmerl2022multilingual}
Katharina Hämmerl, Björn Deiseroth, Patrick Schramowski, Jindřich Libovický, Alexander Fraser, and Kristian Kersting. 2022.
\newblock \href {https://arxiv.org/abs/2203.09904} {Do multilingual language models capture differing moral norms?}
\newblock \emph{Preprint}, arXiv:2203.09904.

\bibitem[{Joshi et~al.(2020)Joshi, Santy, Budhiraja, Bali, and Choudhury}]{joshi-etal-2020-state}
Pratik Joshi, Sebastin Santy, Amar Budhiraja, Kalika Bali, and Monojit Choudhury. 2020.
\newblock \href {https://doi.org/10.18653/v1/2020.acl-main.560} {The state and fate of linguistic diversity and inclusion in the {NLP} world}.
\newblock In \emph{Proceedings of the 58th Annual Meeting of the Association for Computational Linguistics}, pages 6282--6293, Online. Association for Computational Linguistics.

\bibitem[{KhanAcademy(2024)}]{khanmigo}
KhanAcademy. 2024.
\newblock \href {https://www.khanmigo.ai/} {Khanmigo}.
\newblock Website.
\newblock Accessed: 2024-05-21.

\bibitem[{Khanuja et~al.(2021)Khanuja, Bansal, Mehtani, Khosla, Dey, Gopalan, Margam, Aggarwal, Nagipogu, Dave et~al.}]{khanuja2021muril}
Simran Khanuja, Diksha Bansal, Sarvesh Mehtani, Savya Khosla, Atreyee Dey, Balaji Gopalan, Dilip~Kumar Margam, Pooja Aggarwal, Rajiv~Teja Nagipogu, Shachi Dave, et~al. 2021.
\newblock Muril: Multilingual representations for indian languages.
\newblock \emph{arXiv preprint arXiv:2103.10730}.

\bibitem[{Lai and Nissim(2024)}]{lai2024mcot}
Huiyuan Lai and Malvina Nissim. 2024.
\newblock mcot: Multilingual instruction tuning for reasoning consistency in language models.
\newblock \emph{arXiv preprint arXiv:2406.02301}.

\bibitem[{Lewis et~al.(2019)Lewis, Oguz, Rinott, Riedel, and Schwenk}]{DBLP:journals/corr/abs-1910-07475}
Patrick S.~H. Lewis, Barlas Oguz, Ruty Rinott, Sebastian Riedel, and Holger Schwenk. 2019.
\newblock \href {https://arxiv.org/abs/1910.07475} {{MLQA:} evaluating cross-lingual extractive question answering}.
\newblock \emph{CoRR}, abs/1910.07475.

\bibitem[{Li et~al.(2024)Li, Alkhouli, Bonadiman, Pappas, and Mansour}]{li2024eliciting}
Bryan Li, Tamer Alkhouli, Daniele Bonadiman, Nikolaos Pappas, and Saab Mansour. 2024.
\newblock Eliciting better multilingual structured reasoning from llms through code.
\newblock \emph{arXiv preprint arXiv:2403.02567}.

\bibitem[{Lin et~al.(2021)Lin, Mihaylov, Artetxe, Wang, Chen, Simig, Ott, Goyal, Bhosale, Du et~al.}]{lin2021few}
Xi~Victoria Lin, Todor Mihaylov, Mikel Artetxe, Tianlu Wang, Shuohui Chen, Daniel Simig, Myle Ott, Naman Goyal, Shruti Bhosale, Jingfei Du, et~al. 2021.
\newblock Few-shot learning with multilingual language models.
\newblock \emph{arXiv preprint arXiv:2112.10668}.

\bibitem[{Littell et~al.(2017)Littell, Mortensen, Lin, Kairis, Turner, and Levin}]{littell2017uriel}
Patrick Littell, David~R Mortensen, Ke~Lin, Katherine Kairis, Carlisle Turner, and Lori Levin. 2017.
\newblock Uriel and lang2vec: Representing languages as typological, geographical, and phylogenetic vectors.
\newblock In \emph{Proceedings of the 15th Conference of the European Chapter of the Association for Computational Linguistics: Volume 2, Short Papers}, volume~2, pages 8--14.

\bibitem[{Liu et~al.(2023)Liu, Li, Ji, and Lin}]{liu2023towards}
Xiangyan Liu, Rongxue Li, Wei Ji, and Tao Lin. 2023.
\newblock Towards robust multi-modal reasoning via model selection.
\newblock \emph{arXiv preprint arXiv:2310.08446}.

\bibitem[{Liu et~al.(2024)Liu, He, Han, Zhang, Liu, Tian, Zhang, Wang, Gao, Zhong, Pan, Xu, Wu, Liu, Zhang, Zhang, Hu, Zhang, Qiang, Liu, and Ge}]{liu2024understanding}
Yiheng Liu, Hao He, Tianle Han, Xu~Zhang, Mengyuan Liu, Jiaming Tian, Yutong Zhang, Jiaqi Wang, Xiaohui Gao, Tianyang Zhong, Yi~Pan, Shaochen Xu, Zihao Wu, Zhengliang Liu, Xin Zhang, Shu Zhang, Xintao Hu, Tuo Zhang, Ning Qiang, Tianming Liu, and Bao Ge. 2024.
\newblock \href {https://arxiv.org/abs/2401.02038} {Understanding llms: A comprehensive overview from training to inference}.
\newblock \emph{Preprint}, arXiv:2401.02038.

\bibitem[{M365Copilots(2023)}]{m365copilot}
M365Copilots. 2023.
\newblock \href {https://blogs.microsoft.com/blog/2023/03/16/introducing-microsoft-365-copilot-your-copilot-for-work/} {Introducing microsoft 365 copilot – your copilot for work}.
\newblock Microsoft Blog.
\newblock Accessed: 2024-05-21.

\bibitem[{Malaviya et~al.(2017)Malaviya, Neubig, and Littell}]{malaviya17emnlp}
Chaitanya Malaviya, Graham Neubig, and Patrick Littell. 2017.
\newblock Learning language representations for typology prediction.
\newblock In \emph{Conference on Empirical Methods in Natural Language Processing (EMNLP)}, Copenhagen, Denmark.

\bibitem[{Messina et~al.(2023)Messina, Jones, and Poe}]{messina2023prompting}
Cara~Marta Messina, Cherice~Escobar Jones, and Mya Poe. 2023.
\newblock Prompting reflection: Using corpus linguistic methods in the local assessment of reflective writing.
\newblock \emph{Written Communication}, 40(2):620--650.

\bibitem[{Nguyen et~al.(2024)Nguyen, Aljunied, Joty, and Bing}]{prompt1}
Xuan-Phi Nguyen, Sharifah~Mahani Aljunied, Shafiq Joty, and Lidong Bing. 2024.
\newblock \href {https://arxiv.org/abs/2306.11372} {Democratizing llms for low-resource languages by leveraging their english dominant abilities with linguistically-diverse prompts}.
\newblock \emph{Preprint}, arXiv:2306.11372.

\bibitem[{Nie et~al.(2024)Nie, Yuan, Ma, Schmid, F{\"a}rber, Kreuter, and Sch{\"u}tze}]{nie2024decomposed}
Ercong Nie, Shuzhou Yuan, Bolei Ma, Helmut Schmid, Michael F{\"a}rber, Frauke Kreuter, and Hinrich Sch{\"u}tze. 2024.
\newblock Decomposed prompting: Unveiling multilingual linguistic structure knowledge in english-centric large language models.
\newblock \emph{arXiv preprint arXiv:2402.18397}.

\bibitem[{OpenAI(2023)}]{openai2023gpt4}
OpenAI. 2023.
\newblock \href {https://arxiv.org/abs/2303.08774} {Gpt-4 technical report}.
\newblock \emph{Preprint}, arXiv:2303.08774.

\bibitem[{OpenAI(2024)}]{openai2024ada3}
OpenAI. 2024.
\newblock \href {https://openai.com/index/new-embedding-models-and-api-updates/} {New embedding models and api updates}.
\newblock OpenAI Blog.
\newblock Accessed: 2024-05-21.

\bibitem[{Ouyang et~al.(2022)Ouyang, Wu, Jiang, Almeida, Wainwright, Mishkin, Zhang, Agarwal, Slama, Ray, Schulman, Hilton, Kelton, Miller, Simens, Askell, Welinder, Christiano, Leike, and Lowe}]{ouyang2022training}
Long Ouyang, Jeff Wu, Xu~Jiang, Diogo Almeida, Carroll~L. Wainwright, Pamela Mishkin, Chong Zhang, Sandhini Agarwal, Katarina Slama, Alex Ray, John Schulman, Jacob Hilton, Fraser Kelton, Luke Miller, Maddie Simens, Amanda Askell, Peter Welinder, Paul Christiano, Jan Leike, and Ryan Lowe. 2022.
\newblock \href {https://arxiv.org/abs/2203.02155} {Training language models to follow instructions with human feedback}.
\newblock \emph{Preprint}, arXiv:2203.02155.

\bibitem[{Qin et~al.(2024)Qin, Chen, Zhou, Chen, Li, Liao, Li, Che, and Yu}]{qin2024multilingual}
Libo Qin, Qiguang Chen, Yuhang Zhou, Zhi Chen, Yinghui Li, Lizi Liao, Min Li, Wanxiang Che, and Philip~S Yu. 2024.
\newblock Multilingual large language model: A survey of resources, taxonomy and frontiers.
\newblock \emph{arXiv preprint arXiv:2404.04925}.

\bibitem[{Rajpurkar et~al.(2016)Rajpurkar, Zhang, Lopyrev, and Liang}]{DBLP:journals/corr/RajpurkarZLL16}
Pranav Rajpurkar, Jian Zhang, Konstantin Lopyrev, and Percy Liang. 2016.
\newblock \href {https://arxiv.org/abs/1606.05250} {Squad: 100, 000+ questions for machine comprehension of text}.
\newblock \emph{CoRR}, abs/1606.05250.

\bibitem[{Sahoo et~al.(2024)Sahoo, Singh, Saha, Jain, Mondal, and Chadha}]{sahoo2024systematic}
Pranab Sahoo, Ayush~Kumar Singh, Sriparna Saha, Vinija Jain, Samrat Mondal, and Aman Chadha. 2024.
\newblock A systematic survey of prompt engineering in large language models: Techniques and applications.
\newblock \emph{arXiv preprint arXiv:2402.07927}.

\bibitem[{Sarvam(2024)}]{sarvam}
Sarvam. 2024.
\newblock \href {https://analyticsindiamag.com/ai-breakthroughs/sarvam-ai-launches-indias-first-open-source-foundational-model-in-10-indic-languages/} {Sarvam 2b model}.
\newblock Sarvam 2B Model.
\newblock Accessed: 2024-09-15.

\bibitem[{Shen et~al.(2024)Shen, Song, Tan, Li, Lu, and Zhuang}]{shen2024hugginggpt}
Yongliang Shen, Kaitao Song, Xu~Tan, Dongsheng Li, Weiming Lu, and Yueting Zhuang. 2024.
\newblock Hugginggpt: Solving ai tasks with chatgpt and its friends in hugging face.
\newblock \emph{Advances in Neural Information Processing Systems}, 36.

\bibitem[{Shi et~al.(2022)Shi, Suzgun, Freitag, Wang, Srivats, Vosoughi, Chung, Tay, Ruder, Zhou et~al.}]{shi2022language}
Freda Shi, Mirac Suzgun, Markus Freitag, Xuezhi Wang, Suraj Srivats, Soroush Vosoughi, Hyung~Won Chung, Yi~Tay, Sebastian Ruder, Denny Zhou, et~al. 2022.
\newblock Language models are multilingual chain-of-thought reasoners.
\newblock \emph{arXiv preprint arXiv:2210.03057}.

\bibitem[{Shiksha(2024)}]{shiksha}
Microsoft~Research Shiksha. 2024.
\newblock Teachers in india help microsoft research design ai tool for creating great classroom content.
\newblock Microsoft Research Blog.
\newblock Accessed: 2024-05-21.

\bibitem[{Team et~al.(2023)Team, Anil, Borgeaud, Wu, Alayrac, Yu, Soricut, Schalkwyk, Dai, Hauth et~al.}]{team2023gemini}
Gemini Team, Rohan Anil, Sebastian Borgeaud, Yonghui Wu, Jean-Baptiste Alayrac, Jiahui Yu, Radu Soricut, Johan Schalkwyk, Andrew~M Dai, Anja Hauth, et~al. 2023.
\newblock Gemini: a family of highly capable multimodal models.
\newblock \emph{arXiv preprint arXiv:2312.11805}.

\bibitem[{Touvron et~al.(2023)Touvron, Martin, Stone, Albert, Almahairi, Babaei, Bashlykov, Batra, Bhargava, Bhosale et~al.}]{touvron2023llama}
Hugo Touvron, Louis Martin, Kevin Stone, Peter Albert, Amjad Almahairi, Yasmine Babaei, Nikolay Bashlykov, Soumya Batra, Prajjwal Bhargava, Shruti Bhosale, et~al. 2023.
\newblock Llama 2: Open foundation and fine-tuned chat models.
\newblock \emph{arXiv preprint arXiv:2307.09288}.

\bibitem[{Vizcaíno et~al.(2021)Vizcaíno, Saltarin, Belyaev, Lyck, Lasser, and Favaro}]{vizcaino2021convnd}
Josué~Page Vizcaíno, Federico Saltarin, Yury Belyaev, Ruth Lyck, Tobias Lasser, and Paolo Favaro. 2021.
\newblock \href {https://doi.org/10.1109/TCI.2021.3097611} {Learning to reconstruct confocal microscopy stacks from single light field images}.
\newblock \emph{IEEE Transactions on Computational Imaging}, 7:775--788.

\bibitem[{Wang et~al.(2023)Wang, Zheng, Chen, Cai, and Luo}]{wang2023multiple}
Siyuan Wang, Jianming Zheng, Wanyu Chen, Fei Cai, and Xueshan Luo. 2023.
\newblock Multiple: Multilingual prompt learning for relieving semantic confusions in few-shot event detection.
\newblock In \emph{Proceedings of the 32nd ACM International Conference on Information and Knowledge Management}, pages 2676--2685.

\bibitem[{Wang et~al.(2020)Wang, Tsvetkov, and Neubig}]{DBLP:journals/corr/abs-2004-06748}
Xinyi Wang, Yulia Tsvetkov, and Graham Neubig. 2020.
\newblock \href {https://arxiv.org/abs/2004.06748} {Balancing training for multilingual neural machine translation}.
\newblock \emph{CoRR}, abs/2004.06748.

\bibitem[{Wei et~al.(2022)Wei, Wang, Schuurmans, Bosma, Xia, Chi, Le, Zhou et~al.}]{weichain}
Jason Wei, Xuezhi Wang, Dale Schuurmans, Maarten Bosma, Fei Xia, Ed~H Chi, Quoc~V Le, Denny Zhou, et~al. 2022.
\newblock Chain-of-thought prompting elicits reasoning in large language models.
\newblock In \emph{Advances in Neural Information Processing Systems}.

\bibitem[{Wikipedia()}]{iso}
Wikipedia.
\newblock List of iso 639 language codes.
\newblock \url{https://en.wikipedia.org/wiki/List_of_ISO_639_language_codes}.

\bibitem[{Xu et~al.(2024)Xu, Yin, Cai, Yi, Xu, Wang, Wu, Zhao, Yang, Wang et~al.}]{xu2024survey}
Mengwei Xu, Wangsong Yin, Dongqi Cai, Rongjie Yi, Daliang Xu, Qipeng Wang, Bingyang Wu, Yihao Zhao, Chen Yang, Shihe Wang, et~al. 2024.
\newblock A survey of resource-efficient llm and multimodal foundation models.
\newblock \emph{arXiv preprint arXiv:2401.08092}.

\bibitem[{Yang et~al.(2022)Yang, Lin, Yang, Wang, Zhou, and Yang}]{yang2022prompt}
Hao Yang, Junyang Lin, An~Yang, Peng Wang, Chang Zhou, and Hongxia Yang. 2022.
\newblock Prompt tuning for generative multimodal pretrained models.
\newblock \emph{arXiv preprint arXiv:2208.02532}.

\bibitem[{Zhao and Sch{\"u}tze(2021)}]{zhao2021discrete}
Mengjie Zhao and Hinrich Sch{\"u}tze. 2021.
\newblock Discrete and soft prompting for multilingual models.
\newblock \emph{arXiv preprint arXiv:2109.03630}.

\bibitem[{Zohar et~al.(2024)Zohar, Huang, Wang, and Yeung}]{zohar2024lovm}
Orr Zohar, Shih-Cheng Huang, Kuan-Chieh Wang, and Serena Yeung. 2024.
\newblock Lovm: Language-only vision model selection.
\newblock \emph{Advances in Neural Information Processing Systems}, 36.

\bibitem[{Üstün et~al.(2024)Üstün, Aryabumi, Yong, Ko, D'souza, Onilude, Bhandari, Singh, Ooi, Kayid, Vargus, Blunsom, Longpre, Muennighoff, Fadaee, Kreutzer, and Hooker}]{cohereaya}
Ahmet Üstün, Viraat Aryabumi, Zheng-Xin Yong, Wei-Yin Ko, Daniel D'souza, Gbemileke Onilude, Neel Bhandari, Shivalika Singh, Hui-Lee Ooi, Amr Kayid, Freddie Vargus, Phil Blunsom, Shayne Longpre, Niklas Muennighoff, Marzieh Fadaee, Julia Kreutzer, and Sara Hooker. 2024.
\newblock \href {https://arxiv.org/abs/2402.07827} {Aya model: An instruction finetuned open-access multilingual language model}.
\newblock \emph{Preprint}, arXiv:2402.07827.

\end{thebibliography}
\section*{Appendix}

\section{Similar Language Algorithm}
\label{app:sim}
Section 3 introduced various prompt strategies and prompt templates that we have optimized for polyglot LLMs. One of the prompt strategies defined is round-tripping the input in source language through "Similar high-resourced language (Sim)". In this section, we present the algorithm for identifying the right set of similar high-resourced languages for a given source language. For every language, we associate its class attribute between 0-5 based on the classes defined in \cite{joshi-etal-2020-state}. Here, class 5 represents very high-resourced languages like English, whereas 0 represents very low-resourced languages like Gondi, Mundari, etc. We use the language similarity metrics based on  language feature similarities\cite{malaviya17emnlp} captured in lang2vec \cite{littell2017uriel}. We give higher preference to the languages with Latin script since the languages with Latin script have shown better performance on GPTx models\cite{mega}.

\begin{algorithm}
\caption{Get language relevance score based on language similarity distance, the class of related language and whether the related language has Latin script.}
\label{alg:lang_relevance_score}
\SetKwFunction{FMain}{GetRelevanceScore}
\SetKwProg{Fn}{Function}{:}{}
$w_{Latin} \gets 0.9$;

\Fn{\FMain{$d$, $l_{cls}$, $isLatin$}}
  {
  $w \gets 1$; 
  
  \If{$isLatin$}
  {$w \gets w_{Latin}$}
  $score \gets w \times d/l_{cls}$;
  
  return $score$
  }  
\end{algorithm}

\section{Prompt Strategies Results}
\label{app:prompt}
\textbf{Performance of prompts for \tydi.}

\begin{table}[]
\centering
\resizebox{0.5\textwidth}{!}{%
\begin{tabular}{lll|ll}
\hline
     & \multicolumn{2}{l|}{\meta}    & \multicolumn{2}{l}{\gptevalf} \\ \hline
     & \gptt & \turbo & \gptt & \turbo \\ \hline
Mono & \textbf{0.64}        & \textbf{0.64}         & \textbf{0.71}        & \textbf{0.71}         \\
Tans     & 0.49 & 0.51 & 0.61 & 0.63 \\
simi     & 0.47 & 0.47 & 0.58 & 0.58 \\
Aggsrc   & 0.62 & 0.63 & 0.69 & 0.70 \\
aggtrans & 0.49 & 0.52 & 0.60 & 0.63 \\ \hline
\end{tabular}%
}
\caption{Performance of different Prompt strategies for \tydi}
\label{tab:prompt-tydi}
\end{table}

In this section we present the performance of our Prompts on \tydi dataset, We report \meta and \gptevalf for each prompt Averaged across all 9 languages. The numbers are reported for \gptt and \turbo with text-embedding-ada-002 embeddings In Table.\ref{tab:prompt-tydi} we observe similar trends to experiment with \indic, i.e., Each model has different trends across different prompting strategies and the choice of the metrics also favours different model making it difficult to find a suitable choice of prompt for a generalized pipeline.




\begin{algorithm}
\caption{Identifying similar high-resourced languages for a given language}\label{alg:similar_language}
\KwData{Source language $l_{s}$}
\KwResult{A set of similar high-resourced languages $L_{similar}$}
$L_{similar} \gets \emptyset$ \;
$cls_{threshold} \gets 3$ \Comment{Language class threshold}\;
$dist_{threshold} \gets 0.5$ \Comment{Language similarity distance threshold}\;

\For{$l \in L$}{
    \If{$class(l) \geq cls_{threshold}$}{
        $d \gets lang2vec\_distance([\text{syntactic, genetic, geographic}], l, l_{s})$\;
        $RelevanceScore \gets GetRelevanceScore(\text{average}(d), class(l), isLatin(l))$\;
        
        \If{$RelevanceScore \leq dist_{threshold}$}{
            $L_{similar}.add(l)$\;
        }
    }
}
\Return{$L_{similar}$}\;
\end{algorithm}

\textbf{Per language performance for \gptt and \turbo for \indic}

Table. \ref{tab:gpt4-indic-prompt}, \ref{tab:turbo-indic-prompt} presents the performance of \gptt and \turbo respectively with text-embedding-ada-002 embeddings, across all 11 languages and 5 prompts that we propose. Here we observe strong patterns for Agg\_Sim performing the best across majority of the languages ( \(\frac{7}{11}\) for \gptt and \(\frac{5}{11}\) for \turbo), Mono performs better and comes very close to Agg\_sim in these languages. For languages such as "ta", "te" translate is prefered. With the limited languages the variance in the trend is high and a rule based system would fail with inclusion of more languages.  
\begin{table}
\begin{minipage}[t]{0.98\linewidth}
    \centering
    \caption{\gptt on \indic}
    \label{tab:gpt4-indic-prompt}
    \resizebox{1.0\columnwidth}{!}{%
    \begin{tabular}{llllll}
    Lang & Mono          & Translate     & Similar & AggSim                            & AggTrans      \\ \hline
    as   & \textbf{0.58} & 0.33          & 0.33    & \textbf{0.58}                     & 0.32          \\
    bn   & \textbf{0.62} & 0.38          & 0.36    & \textbf{0.62}                     & 0.36          \\
    gu   & \textbf{0.59} & 0.31          & 0.30    & \textbf{0.59}                     & 0.30          \\
    hi   & \textbf{0.67} & 0.54          & 0.42    & {\textbf{0.68}} & 0.51          \\
    kn   & \textbf{0.48} & 0.31          & 0.25    & \textbf{0.48}                     & 0.29          \\
    ml   & \textbf{0.32} & 0.30          & 0.19    & \textbf{0.32}                     & 0.29          \\
    mr   & \textbf{0.58} & 0.33          & 0.30    & \textbf{0.57}                     & 0.32          \\
    or   & \textbf{0.57} & 0.29          & 0.27    & \textbf{0.57}                     & 0.27          \\
    pa   & \textbf{0.61} & 0.46          & 0.43    & 0.60                              & 0.46          \\
    ta   & \textbf{0.31} & \textbf{0.40} & null    & 0.34                              & 0.39          \\
    te   & 0.25          & 0.36          & 0.17    & 0.28                              & \textbf{0.37} \\ \hline
    AVG  & \textbf{0.51} & 0.36          & 0.30    & \textbf{0.51}                     & 0.35          \\ \hline
    \end{tabular}%
    }
    
\end{minipage}
~
\begin{minipage}[t]{0.98\linewidth}
    \caption{\turbo on \indic}
    \label{tab:turbo-indic-prompt}
    \centering
    \resizebox{1.0\columnwidth}{!}{%
    \begin{tabular}{lrrlll}
    Lang & {Mono}          & {Translate}     & Similar                  & AggSim                            & AggTrans                 \\ \hline
    as   & {\textbf{0.40}} & {0.34}          & 0.33                     & \textbf{0.45}                     & 0.37                     \\
    bn   & {\textbf{0.54}} & {0.39}          & {0.34} & \textbf{0.54}                     & 0.46                     \\
    gu   & \textbf{0.48}                     & 0.32                              & {0.30} & {\textbf{0.49}} & {0.33} \\
    hi   & \textbf{0.63}                     & 0.52                              & 0.38                     & \textbf{0.64}                     & 0.52                     \\
    kn   & \textbf{0.47}                     & 0.32                              & {0.21} & \textbf{0.46}                     & {0.31} \\
    ml   & \textbf{0.23}                     & \textbf{0.32}                     & 0.13                     & \textbf{0.26}                     & 0.31                     \\
    mr   & \textbf{0.48}                     & 0.34                              & 0.30                     & \textbf{0.47}                     & 0.36                     \\
    or   & \textbf{0.40}                     & 0.29                              & {0.27} & \textbf{0.39}                     & 0.32                     \\
    pa   & \textbf{0.54}                     & 0.46                              & {0.40} & \textbf{0.54}                     & 0.44                     \\
    ta   & {\textbf{0.31}} & {\textbf{0.40}} & null                     & 0.24                              & \textbf{0.39}            \\
    te   & {0.25}          & {\textbf{0.36}} & 0.17                     & 0.25                              & \textbf{0.34}            \\ \hline
    AVG  & {\textbf{0.43}} & {0.37}          & 0.28                     & \textbf{0.43}                     & 0.38                     \\ \hline
    \end{tabular}%
    }
    
\end{minipage}
\end{table}


\textbf{Per language performance for \gptt and \turbo for \tydi}
In Table. \ref{tab:gptt-tydi-prompt}, \ref{tab:turbo-tydi-prompt} performance of \gptt and \turbo along with text-embedding-ada-002 embeddings are presented across all 9 languages and 5 proposed prompts. Here contrary to \indic experiments Mono is preferred over Agg\_sim making a significant change in distribution. The optimal prompt doesn't depend only on the language or model but also on the distribution of the question, this statement is supported by the fact \tydi and \indic share 2 languages "bn" and "te", while in \indic Agg\_sim was prefered for "bn" and Translate for "te" it has completely shifted to Mono for both "bn" and "te" in \tydi. Hence prompt selection is depends on the language and also the distribution of the dataset or sample.

\begin{table}[]
\begin{minipage}[t]{0.98\linewidth}
    \centering
    \caption{\gptt on \tydi}
    \label{tab:gptt-tydi-prompt}
    \resizebox{1.0\columnwidth}{!}{%
    \begin{tabular}{llllll}
    Lang & Mono          & Translate     & Similar & AggSim        & AggTrans \\ \hline
    ar   & \textbf{0.50} & 0.43          & null    & \textbf{0.50} & 0.40     \\
    bn   & \textbf{0.69} & 0.46          & 0.43    & \textbf{0.69} & 0.43     \\
    en   & \textbf{0.65} & null          & 0.60    & 0.62 & 0.58     \\
    fi   & \textbf{0.63} & 0.49          & 0.48    & 0.59 & 0.49     \\
    id   & \textbf{0.66} & 0.58          & 0.53    & 0.63 & 0.54     \\
    ko   & \textbf{0.64} & {0.48}        & 0.43    & {0.63} & 0.47     \\
    ru   & \textbf{0.51} & 0.45          & 0.46    & {0.50} & 0.44     \\
    sw   & \textbf{0.80} & 0.63          & null    & {0.78} & 0.65     \\
    te   & \textbf{0.67} & 0.42          & 0.39    & {0.66} & 0.43     \\ \hline
    AVG  & \textbf{0.64} & 0.49          & 0.47    & {0.62} & 0.49     \\ \hline
    \end{tabular}%
    }
\end{minipage}
~
\begin{minipage}[t]{0.98\linewidth}
    \centering
    \caption{\turbo on \tydi}
    \label{tab:turbo-tydi-prompt}
    \resizebox{1.0\columnwidth}{!}{%
    \begin{tabular}{llllll}
    Lang & Mono          & Translate & Similar & AggSim & AggTrans \\ \hline
    ar   & 0.53          & 0.45      & null    & 0.52   & 0.42     \\
    bn   & 0.65          & 0.46      & 0.41    & 0.65   & 0.48     \\
    en   & 0.66          & null      & 0.61    & 0.65   & 0.64     \\
    fi   & 0.68          & 0.53      & 0.50    & 0.65   & 0.54     \\
    id   & 0.67          & 0.61      & 0.53    & 0.66   & 0.59     \\
    ko   & 0.65          & 0.49      & 0.46    & 0.66   & 0.51     \\
    ru   & 0.52          & 0.46      & 0.45    & 0.51   & 0.44     \\
    sw   & 0.76          & 0.64      & null    & 0.74   & 0.64     \\
    te   & 0.66          & 0.44      & 0.36    & 0.67   & 0.43     \\ \hline
    AVG  & \textbf{0.64} & 0.51      & 0.47    & 0.63   & 0.52     \\ \hline
    \end{tabular}%
    }
\end{minipage}
\end{table}


\section{Hybrid approach}
\label{app:hybrid}

In this section we evaluate the performance of our Hybrid Approach across text-ada-002-embedding, Adav3, XLMRXXL and Cohere embed\_multilingual\_v3. We use \tydi as the dataset and average the \meta and \gptevalf across all 9 languages and all 5 prompts. In Table. \ref{tab:hybrid-perf-tydi} we present the values for both \gptt and \turbo, while the trend is completely different to that of \indic which could be primarily attributed to the languages typology and derivations.

\begin{table*}[]
\centering
\resizebox{0.85\textwidth}{!}{%
\begin{tabular}{llllll}
\hline
Metrics & Models                & Ada           & Adav3 & XLMRXXL & Cohere        \\ \hline
\multirow{2}{*}{\meta}     & \gptt & \textbf{0.64} & \textbf{0.64} & 0.60 & \textbf{0.61} \\
        & \turbo & \textbf{0.64} & 0.60  & 0.57    & \textbf{0.59} \\ \hline
\multirow{2}{*}{\gptevalf} & \gptt & \textbf{0.71} & \textbf{0.71} & 0.65 & \textbf{0.68} \\
        & \turbo & \textbf{0.71} & 0.66  & 0.63    & \textbf{0.66} \\ \hline
\end{tabular}%
}
\caption{Hybrid approach performance - \tydi.}
\label{tab:hybrid-perf-tydi}
\end{table*}







    









\section{Detailed Training Procedure \& Implemenation Details}\label{app:training}\label{app:hybrid_arch}

The algorithm employs separate strategies for inference and training tailored to different operational conditions. During inference, the algorithm selects the optimal configuration based on F1 score predictions from task and configuration embeddings as described in Algorithm \ref{alg:learning}. In the Offline Setting, configuration selection is deterministic, using an argmax function for precise, data-rich environments. Conversely, the Online Setting uses a probabilistic softmax function to adapt to data-scarce situations, enabling dynamic exploration and refinement of configurations.

For training, the offline mode applies a Mean Squared Error (MSE) loss across all configurations, ensuring comprehensive learning. In contrast, the online mode implements a sparse MSE loss, updating only the evaluated configurations through a masking technique. This approach reduces computational load and accelerates adaptation to new data, optimizing performance in real-time applications as outlined in Algorithm \ref{alg:learning}.

The sparse MSE Loss employs a Mask \(\mathcal{M}\) which is defined as,
Let \(\mathbf{C}\) be a tensor of order \(m\) with dimensions \(n_1 \times n_2 \times \dots \times n_m\). Suppose \(\hat{c} = C_{i_1, i_2, \dots, i_m}\) is a selected element from \(\mathbf{C}\), where \((i_1, i_2, \dots, i_m)\) are the indices of \(\hat{c}\) in \(\mathbf{C}\). Define the tensor \(\mathcal{M}\) as follows:

\begin{equation}
\footnotesize
\mathcal{M}_{j_1, j_2, \dots, j_m} = 
\begin{cases} 
1 & \text{if } (j_1, j_2, \dots, j_m) = (i_1, i_2, \dots, i_m) \\
0 & \text{otherwise}
\end{cases}
\label{eq:mask_matrix}
\end{equation}

\begin{algorithm}
\caption{Learning Strategy Algorithm for Inference and Training}\label{alg:learning}
\DontPrintSemicolon  

\KwData{Task descriptions \(\mathcal{T}\), configuration options \(\mathcal{C}_i\)}
\KwResult{Optimal configuration \(\hat{c}\) and its corresponding F1 score}

\(\mathcal{B}\) - LLaMa-2-70B backbone for embedding generation\;
\(\mathcal{H}\) - Conv-ND layers for F1 score prediction\;
\(e\) - embedding projection size, \(e = 8192\)\;
\(m\) - number of parameters, \(m = 3\) (e.g., language model, embedding model, prompt strategies)\;
\(\mathcal{R}^{n_1 \times n_2 \times \dots \times n_m}\) - size of the N-dimensional array for configurations\;
\(bs\) - Batch size of Task Definitions.

\For{\(\mathcal{T}_{j} \leftarrow \{\mathcal{T}_{0},...\mathcal{T}_{bs} \} \)}{
    \;
    
    \(\mathcal{E}_{Tj} \leftarrow \mathcal{B}(\mathcal{T}_{j})\) ;
    \(\mathcal{E}_{Ci} \leftarrow \mathcal{B}(\mathcal{C}_{i})\) \\
    \(\mathcal{E}_{j} \leftarrow \text{Concatenate}(\mathcal{E}_{Tj}, \mathcal{E}_{Ci})\) 
    
    \(\hat{y} \leftarrow \mathcal{H}(\mathcal{E}_{j})\) 
    
    \;
    
    {\color{blue}\tcc{Inference for selecting configuration}}

    \uIf{Offline Setting}{
    \(\hat{c} \leftarrow \arg\max(\hat{y})\)
    }\uElseIf{Online Setting}{
    \(\hat{c} \sim \text{Softmax}(\hat{y})\)
    }
    
    \;
    
    {\color{blue}\tcc{Training to update \(\mathcal{H}\) \& \(\mathcal{B}\) }}

    \uIf{Offline Setting}{
        \(y \leftarrow \text{Ground truth F1 scores }  \forall \mathcal{C}_{i} \)\;
        \(Loss_{\text{off}} \leftarrow \text{MSE}(\hat{y}, y)\)
    }
    \uElseIf{Online Setting}{
        \(y_{\text{sparse}} \leftarrow \text{Ground truth F1 score for } \hat{c} \)\;
        \(\mathcal{M} \leftarrow \text{Mask matrix using eq. \ref{eq:mask_matrix} }\)\;
        \(Loss_{\text{on}} \leftarrow \text{MSE}(\mathcal{M} \odot \hat{y}, y_{\text{sparse}})\)
    }

    \;
    
    Update \(\mathcal{H}\) \& \(\mathcal{B}\) using \(Loss\)
}
\end{algorithm}

\subsection{Implementation Details}
\label{sec:imple}
In this work, we use Azure OpenAI models~\cite{azureopenai} for all our LLM and embedding models including \gptt, \turbo and \mix. For the given configurations of base LLM models, embeddings and prompt strategies, the training needs to be performed only once and can be shared with different multilingual applications and use-cases. For the learning model, we train the Llama model on GPU with A100 80 GB, CPU	with 96 cpu cores at 2.2GHz and 1024 GB RAM. The duration of training 100 offline epochs is 1.42 Hrs. The duration of training 25 online epochs is 0.74 Hrs. The inference and evaluation is dependent on the rate limits imposed by Azure OpenAI APIs~\cite{azureopenai}.
\textbf{Model Version}: For LLMs we use \gptt - 0125-preview, \turbo - 0125, Mixtral - Mixtral-8x7B-Instruct-v0.1; For Embeddings we use ada - text-ada-002-embedding, ada3 - text-ada-003-embedding, XLMR-XXL - facebook/xlm-roberta-xxl and Cohere - embed\_multilingual\_v3;.

\section{\gpteval Setup and details}
\label{sec:gptanno}

\subsection{Human Annotation Task Details}
\label{subsec:human_annotation}

We build a simple human annotation interface using Streamlit\footnote{https://streamlit.io/} where the context, the question related to the context, and the ground truth answer for each record are fetched from the IndicQA dataset\cite{ai4bharat2022indicqa}. In this evaluation task, the annotators are first presented with a passage that acts as the context required to answer the question which is shown along with the ground truth answer. The annotators are then asked to evaluate the answers generated by the LLM using different strategies based on the ground truth answer provided, by answering one of the following options: "Yes", "No" or "Partial".  Here is the instruction provided to the annotators. 

\fbox{\begin{minipage}{18em}
First, select your language and go through the context under the title "Context GT" once. Then, look at the question and try to answer this question and compare it with the ground truth answer. Next, for all the available answers, choose:
\begin{enumerate}
    \item "Yes" if the answer is absolutely correct(minor punctuation errors are allowed)
    \item "Partial" if the answer captures some part of the core answer, but has grammatical mistakes or minor errors(spelling, etc.) that make the answer partially correct.
    \item "No" if the answer is completely wrong
\end{enumerate}
\end{minipage}}

Based on the human annotations for each question, we then recompute the F1 score. The updated F1 scores are calculated using Algorithm \ref{alg:eval}, where $evals$ contains evaluations for all the strategies annotated by the human annotator.


\begin{algorithm}
\caption{Evaluation Algorithm when using Human Annotator or \gpteval}\label{alg:eval}
\KwData{$ground\_truth$, $gpt\_answers$, $evals$}
\KwResult{$eval\_scores$}
$eval\_scores \gets []$;
$valid\_answers \gets []$\;
$evals = get\_eval(gpt\_answers)$;
$valid\_answers.append(ground\_truth)$;
\For{$i\gets0$ \KwTo $len(gpt\_answers)$} {
    \If{$evals[i] = "Yes"$} {
    $valid\_answers.append(gpt\_answers[i])$;
    }
}
\For{$i\gets0$ \KwTo $len(gpt\_answers)$} {
    $eval\_f1.append(compute\_score$\\$(gpt\_answers[i], valid\_answers))$;
}
\end{algorithm}

\subsection{GPT Eval process}

In Section \ref{sec:limitations}, we introduced \gpteval, where GPT models perform the evaluation of the answer generated when compared to the ground truth. Similar to the human evaluation task described in the previous subsection \ref{subsec:human_annotation}, the \gpteval is tasked to evaluate the LLM responses based on the available ground truth for the given record. The prompt below is used for \turbo in order to evaluate the answers.

\fbox{
\begin{minipage}{18em}
You are a multilingual evaluation assistant. Users will send in a query, context text, the correct answer for the query based on the context text, and also an answer that needs to be evaluated. You will evaluate the answer based on the context text and the correct answer that the user has sent and respond with Yes, No, or Partial based on the below evaluation instructions. Instructions: 1. Yes if the answer is absolutely correct. 2. Partial if the answer captures some part of the correct answer, but has minor errors like grammatical or spelling mistakes, etc. 3.No if the answer is completely wrong.
\end{minipage}}

The updated F1 Scores for each of the strategy is calculated using Algorithm \ref{alg:eval} where $evals$ contains "Yes", "No" or "Partial" evaluations as judged by the \gpteval.
\end{document}